\documentclass[11pt,a4paper]{article}
\usepackage[hyperref]{acl2021}
\usepackage{times}
\usepackage{latexsym}

\usepackage{url}
\usepackage{bm}
\usepackage{amsfonts}
\usepackage{float}
\usepackage{graphicx}
\usepackage{soul}
\usepackage{url}
\usepackage{booktabs}
\usepackage{algorithm}
\usepackage{algorithmic}
\usepackage{amsmath}
\usepackage{multirow}

\usepackage{color}

\usepackage{microtype}

\aclfinalcopy 
\def\aclpaperid{3738} 

\title{Topic-Aware Evidence Reasoning and Stance-Aware Aggregation for Fact Verification}

\author{Jiasheng Si$^{\dag}$ \ \ \ \   Deyu Zhou$^{\dag}\thanks{corresponding author}$ \ \ \ \  Tongzhe Li$^{\dag}$ \ \ \ \  Xingyu Shi$^{\dag}$ \ \ \ \  Yulan He$^{\S}$ \\
$^{\dag}$ School of Computer Science and Engineering, Key Laboratory of Computer Network\\
	and Information Integration, Ministry of Education, Southeast University, China \\
$\S$ Department of Computer Science, University of Warwick, UK \\
\texttt{\{jasenchn, d.zhou, 220184611, xyu-shi\}@seu.edu.cn}, \\
\texttt{yulan.he@warwick.ac.uk}}

\date{}

\begin{document}
\maketitle
\begin{abstract}
    Fact verification is a challenging task that requires simultaneously reasoning and aggregating over multiple retrieved pieces of evidence to evaluate the truthfulness of a claim.
    Existing approaches typically
    (i) explore the semantic interaction between the claim and evidence at different granularity levels but fail to capture their topical consistency during the reasoning process,
    which we believe is crucial for verification;
    (ii) aggregate multiple
    pieces of evidence equally without considering
    their implicit stances to the claim, thereby introducing spurious information.
    To alleviate the above issues,
    we propose a novel topic-aware evidence reasoning and stance-aware aggregation model for more accurate fact verification, with the following four key properties:
    1) checking topical consistency between the claim and evidence;
    2) maintaining topical coherence among multiple pieces of evidence;
    3) ensuring semantic similarity between the global topic information and the semantic representation of evidence;
    4) aggregating evidence based on their implicit stances to the claim.
    Extensive experiments conducted on the two benchmark datasets demonstrate the superiority of the proposed model over several state-of-the-art approaches for fact verification.
    The source code can be obtained from \url{https://github.com/jasenchn/TARSA}.

\end{abstract}

\section{Introduction}
\label{introduction}
The Internet breaks the physical distance barrier among individuals to allow them to share data and information online.
However, it can also be used by people with malicious purposes to
disseminate misinformation or fake news.
Such misinformation may cause ethnics conflicts, financial losses and political unrest,
which has become one of the greatest threats to the public
~\citep{DBLP:conf/kdd/ZafaraniZSL19, DBLP:conf/wsdm/ZhouZSL19}.
Moreover, as shown in ~\citet{vosoughi2018spread},
compared with truth,
misinformation diffuses significantly farther, faster, and deeper in all genres.
Therefore, there is an urgent need for quickly identifying
the misinformation spread on the web.
To solve this problem,
we focus on the fact verification task~\citep{DBLP:conf/naacl/ThorneVCM18},
which aims to automatically evaluate the veracity of a given claim based on the textual evidence retrieved from external sources.

\begin{figure*}[!h]
    \centering
    \includegraphics[width=0.9\linewidth]{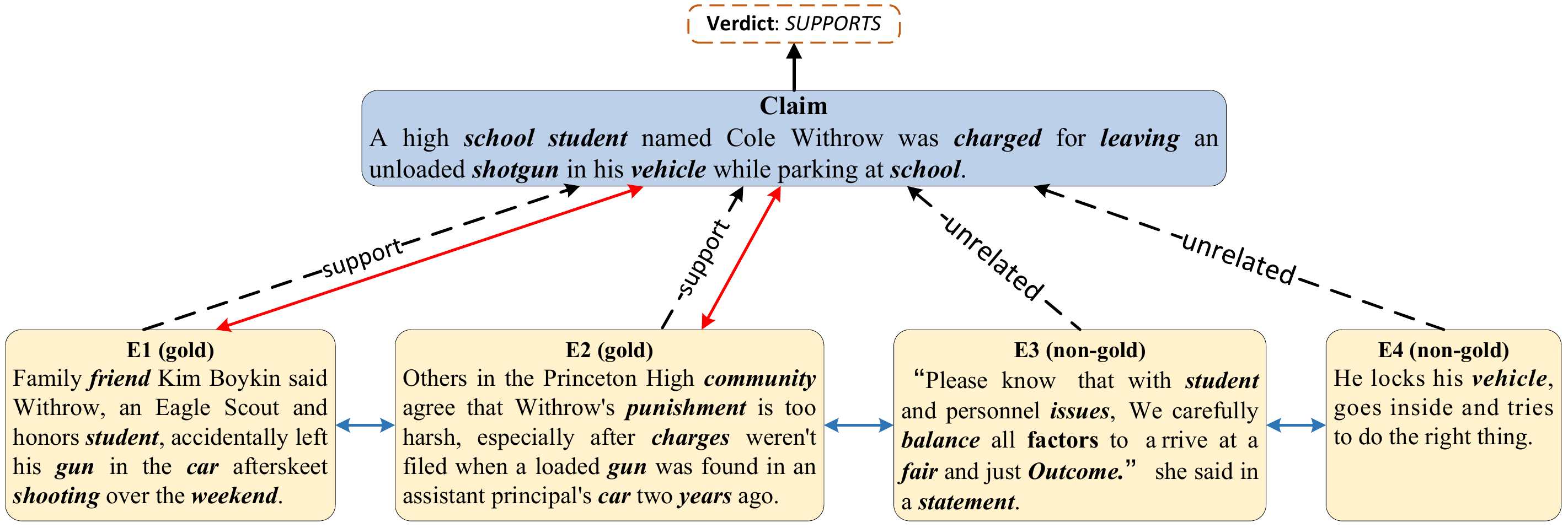}
    \caption{An example of fact verification.
        The bold italic words are topic words extracted by latent Dirichlet allocation (LDA).
        The red solid line denotes the topical consistency between the claim and evidence.
        The black dotted line denotes the implicit stance of evidence towards the claim.
        The blue solid line denotes the topical coherence among evidence.}
    \label{fig:example}
\end{figure*}

Recent approaches for fact verification are dominated by natural language inference models~\citep{DBLP:conf/emnlp/AngeliM14} or 
textual entailment recognition models~\citep{DBLP:conf/acl/MaGJW19},
where the truthfulness of a claim is verified via reasoning and aggregating over multiple pieces of retrieved evidence.
In general,
existing models follow an architecture with two main sub-modules:
the semantic interaction module and the entailment-based aggregation module~\citep{DBLP:conf/coling/HanselowskiSSCC18,DBLP:conf/aaai/NieCB19,DBLP:conf/ecir/SoleimaniMW20,DBLP:conf/acl/LiuXSL20}.
The semantic interaction module attempts to grasp the rich semantic-level interactions among multiple pieces of evidence at the sentence-level~\citep{DBLP:conf/acl/MaGJW19,DBLP:conf/acl/ZhouHYLWLS19,DBLP:conf/emnlp/SubramanianL20} or the semantic roles-level~\citep{DBLP:conf/acl/ZhongXTXDZWY20}.
The entailment-based aggregation module aims to filter out irrelevant information to capture the salient information related to the claim by aggregating the semantic information coherently.

However, the aforementioned approaches
typically learn the representation of each evidence-claim pair from the semantic perspective
such as obtaining
the semantic representation of each evidence-claim pair through pre-trained language models~\citep{DBLP:conf/naacl/DevlinCLT19} or graph-based models~\citep{DBLP:conf/iclr/VelickovicCCRLB18},
which largely overlooked the topical consistency between
claim and evidence.
For example in Figure~\ref{fig:example},
given the claim ``\emph{A high school student named Cole Withrow was charged for leaving an unloaded shotgun in his vehicle while parking at school}'' and the retrieved evidence sentences (i.e., $E1$-$E4$), we would expect a fact checking model to automatically filter evidence which is topically-unrelated to the claim such as $E3$ and $E4$ and only relies on the evidence which is topically-consistent with the claim such as $E1$ and $E2$ for veracity assessment of the claim.
In addition, we also expect the topical coherence of multiple pieces of supporting evidence such as $E1$ and $E2$.
Furthermore,
in previous approaches, the learned representations of multiple pieces of evidence
are aggregated via element-wise max pooling or simple dot-product attention, which inevitably fails to capture the implicit stances of evidence toward the claim (e.g., $E1$ and $E2$ \emph{support} the claim implicitly, $E3$ and $E4$ are \emph{unrelated} to the claim) and leads to the combination of irrelevant information with relevant one.

To address these problems,
in this paper,
we propose a novel neural structure reasoning model for fact verification,
named TARSA (\underline{T}opic-\underline{A}ware Evidence \underline{R}easoning and \underline{S}tance-Aware \underline{A}ggregation Model).
A coherence-based topic attention is developed to model the topical consistency between a claim and each piece of evidence and the topical coherence among evidence built on the sentence-level topical representations.
In addition, a semantic-topic co-attention is created to measure
the coherence between the global topical information and the semantic representation of the claim and evidence.
Moreover, the capsule network is incorporated to model the implicit stances of evidence toward the claim by the dynamic routing mechanism.

The main contributions are listed as follows:
\begin{itemize}
    \item  We propose a novel topic-aware evidence reasoning and stance-aware aggregation approach,
          which is, to our best knowledge, the first attempt of jointly exploiting semantic interaction and topical consistency to learn latent evidence representation for fact verification.
    \item  We incorporate the capsule network structure into our proposed model to capture the
          implicit stance relations between the claim and the evidence.
    \item  We conduct extensive experiments on the two benchmark datasets to demonstrate the effectiveness of TARSA for fact verification.

\end{itemize}

\section{Related Work}

In general, fact verification is a task to assess the authenticity of a claim backed by a validated corpus of documents,
which can be divided into two stages: fact extraction and claim verification~\citep{DBLP:journals/csur/ZhouZ20}.
Fact extraction can be further split into the document retrieval phase and the evidence selection phase to shrink the search space of evidence~\citep{DBLP:conf/naacl/ThorneVCM18}.
In the document retrieval phase,
researchers typically reuse the top performing approaches in the FEVER1.0 challenge to extract the documents with high relevance for a given claim~\citep{DBLP:journals/corr/abs-1809-01479,yoneda2018ucl,DBLP:conf/aaai/NieCB19}.
In the evidence selection phase, to select relevant sentences, researchers generally train the classification models or rank models based on the similarity between the claim and each sentence from the retrieved documents~\citep{DBLP:conf/acl/ChenZLWJI17,stammbach2019team,DBLP:conf/ecir/SoleimaniMW20,DBLP:conf/emnlp/WaddenLLWZCH20,DBLP:conf/acl/ZhongXTXDZWY20,DBLP:conf/acl/ZhouHYLWLS19}.

Many fact verification approaches focus on the claim verification stage,
which can be addressed by natural language inference methods~\citep{DBLP:conf/emnlp/ParikhT0U16,DBLP:conf/naacl/GhaeiniHDLLQLPF18,luken-etal-2018-qed2}.
Typically, these approaches contain the representation learning process and evidence aggregation process.
\citet{DBLP:journals/corr/abs-1809-01479} and~\citet{DBLP:conf/aaai/NieCB19} concatenate all pieces of evidence as input and use the max pooling to aggregate the information for claim verification via the enhanced sequential inference model (ESIM)~\citep{DBLP:conf/acl/ChenZLWJI17}.
In a similar vein,
\citet{DBLP:conf/emnlp/0001R18} incorporate the identification of evidence to further improve claim verification using ESIM with different granularity levels.
\citet{DBLP:conf/acl/MaGJW19} leverage the co-attention mechanism between claim and evidence to generate claim-specific evidence representations which are used to infer the claim. 

Benefiting from the development of pre-trained language models,
\citet{DBLP:conf/acl/ZhouHYLWLS19} are the first to learn evidence representations by BERT~\citep{DBLP:conf/naacl/DevlinCLT19}, 
which are subsequently used in a constructed evidence graph for claim inference by aggregating all claim-evidence pairs.
\citet{DBLP:conf/acl/ZhongXTXDZWY20} further establish a semantic-based graph for representation and aggregation with XLNet~\citep{DBLP:conf/nips/YangDYCSL19}.
\citet{DBLP:conf/acl/LiuXSL20} incorporate two sets of kernels into a sentence-level graph to learn a more fine-grained evidence representations.
\citet{DBLP:conf/emnlp/SubramanianL20} further incorporate evidence set retrieval and hierarchical attention sum block to improve the performance of claim verification.

Different from all previous approaches,
our work for the first time handles the fact verification task by considering the topical consistency and the semantic interactions between claim and evidence.
Moreover,
we employ the capsule network to model the implicit stance relations of evidence toward the claim.

\section{Method}
\begin{figure*}[!h]
    \centering
    \includegraphics[width=0.8\linewidth]{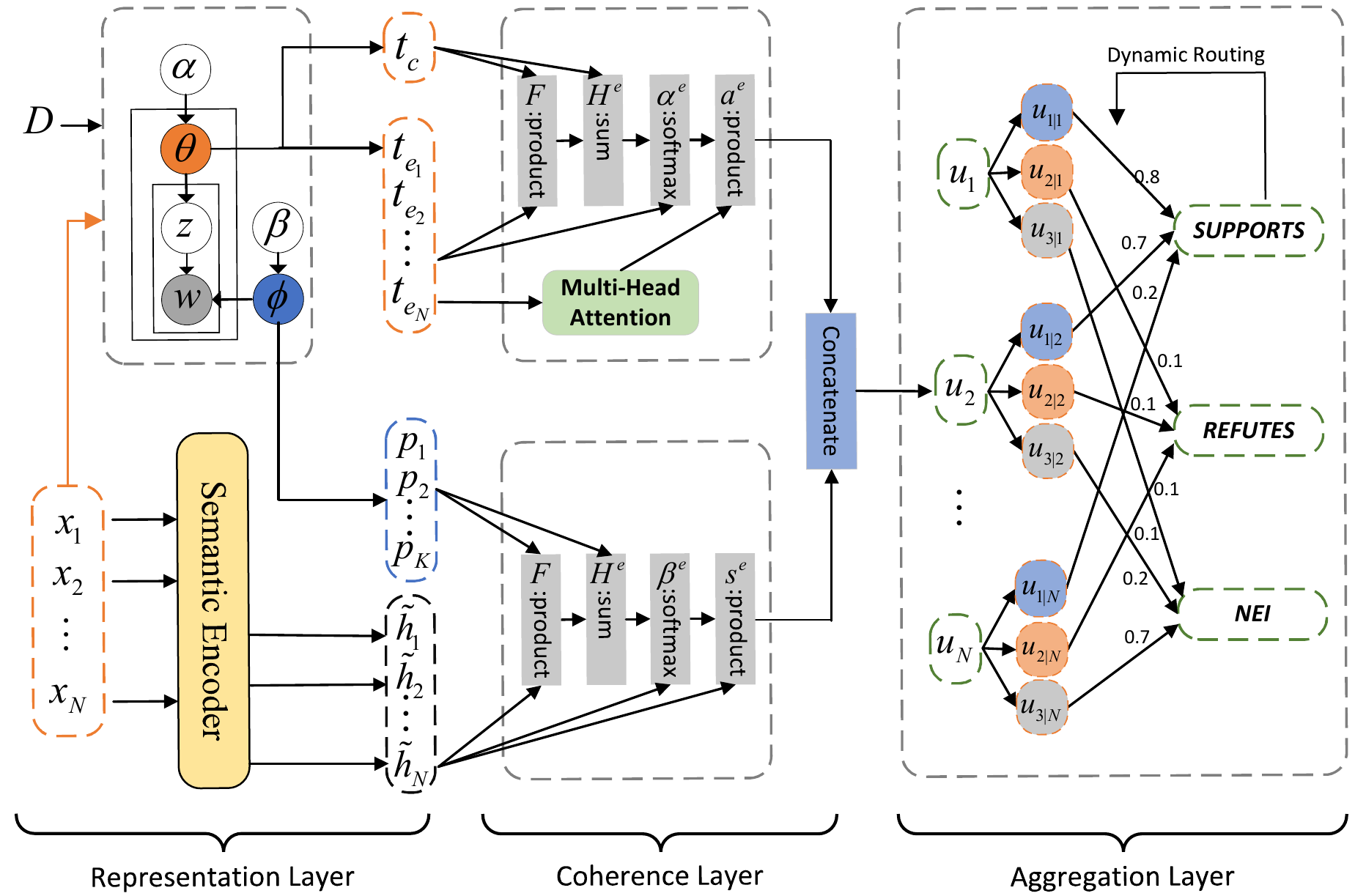}
    \caption{The overview of the architecture of our Topic-Aware Evidence Reasoning and Stance-Aware Aggregation model (TARSA)}
    \label{fig:structure}
\end{figure*}

In this section, we present an overview of the architecture of the proposed framework TARSA for fact verification.
As shown in Figure~\ref{fig:structure},
our approach consists of three main layers:
1) the representation layer to embed claim and evidence into three types of representations by a semantic encoder and a topic encoder;
2) the coherence layer to incorporate the topic information into our model by two attention components;
3) the aggregation layer to model the implicit stances of evidence toward claim using the capsule network.

\subsection{Representation Layer}
This section describes how TARSA extracts semantic representations, sentence-level topic representations,
and global topic information through a semantic encoder and a topic encoder separately.

\paragraph{Semantic Encoder}
The semantic encoder in TARSA is a vanilla transformer~\citep{DBLP:conf/nips/VaswaniSPUJGKP17} with the eXtra hop attention~\citep{DBLP:conf/iclr/ZhaoXRSBT20}.
For each claim $c$ paired with $N$ pieces of retrieved evidence sentences $E=\{e_{1}, e_{2},\cdots,e_{N}\}$,
TARSA constructs the evidence graph by treating each evidence-claim pair $x_{i}=(e_{i}, c)$ as a node (i.e., $x_{i}=\big[[CLS];e_{i};[SEP];c;[SEP]\big]$) and build a fully-connected evidence graph $G$.
We also add a \emph{self-loop} to every node to perform message propagation from itself.

Specifically,
we first apply the vanilla transformer on each node to generate the claim-dependent evidence representation using the input $x_i$,
\begin{equation}
    \label{eq:in-sequence attention}
    \bm{h}_{i} = Transformer(x_i)
\end{equation}
where $i$ denotes the $i$-th node  in $G$.
We treat the first token representation $\bm{h}_{i,0}$ as the local context of node $i$.

Then the eXtra hop attention takes the $[CLS]$ token in each node as a ``hub token",
which is to attend on hub tokens of all other connected nodes to learn the global context.
One layer of eXtra hop attention can be viewed as a single-hop message propagation among all the nodes along the edges,
\begin{equation}
    \label{eq:extra hop attention}
    \hat{\bm{h}}_{i,0} = \sum_{j; e_{i,j}=1}{softmax_{j}(\frac{\hat{\bm{q}}_{i,0}^T \cdot \hat{\bm{k}}_{j,0}}{\sqrt{d_k}})\cdot \hat{\bm{\nu}}_{j, 0}}
\end{equation}
where $e_{i,j}=1$ denotes that there is an edge between the node $i$ and the node $j$,
$\hat{\bm{q}}_{i,0}$ denotes the \emph{query} vector of the $[CLS]$ token of node $i$,
$\hat{\bm{k}}_{j,0}$ and $\hat{\bm{\nu}}_{j, 0}$ denote the \emph{key} vector and the $value$ vector of the $[CLS]$ token of node $j$, respectively,
and $\sqrt{d_k}$ denotes the scaling factor.

The local context and the global context are concatenated to learn the semantic representation of all the nodes:
\begin{equation}
    \begin{aligned}
        \label{eq:extra hop attention}
         & \tilde{\bm{h}}_{i,0}=Linear([\bm{h}_{i,0}; \hat{\bm{h}}_{i,0}]),   \\
         & \tilde{\bm{h}}_{i,\tau}=\bm{h}_{i,\tau}; \forall \tau \neq 0.
    \end{aligned}
\end{equation}

By stacking $L$ layers of the transformer with the eXtra hop attention which takes the semantic representation of the previous layer as input,
we learn the semantic representation of evidence $\bm{H}=[\tilde{\bm{h}}_1,\tilde{\bm{h}}_2,\cdots,\tilde{\bm{h}}_N]\in \mathbb{R}^{N\times d}$ from the graph $G$.

\paragraph{Topic Encoder}
We extract topics in the following two forms via latent Dirichlet allocation (LDA)~\citep{DBLP:journals/jmlr/BleiNJ03}:

\noindent\underline{Sentence-level topic representation:} Given a claim $c$ and $N$ pieces of the retrieved evidence $E$,
we extract latent topic distribution $\bm{t}\in\mathbb{R}^K$ for each sentence as the sentence-level topic representation,
where $K$ is the number of topics.
More concretely, we denote $\bm{t}_{c}\in\mathbb{R}^K$ for claim $c$ and $\bm{t}_{e_{i}}\in\mathbb{R}^K$ for evidence $e_{i}$.
Each scalar value $t^k$ denotes the contribution of topic $k$ in representing the claim or evidence.

\noindent\underline{Global topic information:}
We extract global topic information $\bm{P}=[\bm{p}_1,\bm{p}_2,\cdots,\bm{p}_K]\in\mathbb{R}^{K\times V}$ from the topic-word distribution
by treating each sentence (i.e., claim or evidence) in corpus $D$ as a document,
where $V$ denotes the vocabulary size.

\subsection{Coherence Layer}
This section describes how to incorporate the topic information into our model with two attention components.

\paragraph{Coherence-based Topic Attention}
\label{topic information}

Based on the observation as illustrated in Figure~\ref{fig:example}, we assume that \emph{given a claim, the sentences used as evidence should be topically coherent with each other} and \emph{the claim should be topically consistent with the relevant evidence}.
Therefore, two kinds of topical relationship are considered:
1) topical coherence among multiple pieces of evidence ($TC_{ee}$);
2) topical consistency between the claim and each evidence ($TC_{ce}$).

Specifically,
to incorporate the topical coherence among multiple pieces of evidence into our model,
we disregard the order of evidence and treat each evidence independently.
Then we utilize the multi-head attention~\citep{DBLP:conf/nips/VaswaniSPUJGKP17} without position embedding to generate the new topic representation of evidence $\hat{\bm{t}}_e$
based on the sentence-level topic representation $\bm{t}_e\in\mathbb{R}^{N\times K}$ of the retrieved evidence for a given claim.
\begin{equation}
    \label{local coherence}
    \bm{\hat{t}}_e=multihead(\bm{t}_{e})
\end{equation}

Moreover,
we utilize the co-attention mechanism~\citep{DBLP:conf/emnlp/ChenL20} to weigh each evidence based on the topic consistency between the claim and the evidence.
Given the sentence-level topic representation $\bm{t}_c$ for claim and $\bm{t}_e$ for the corresponding evidence,
the co-attention attends to the claim and the evidence simultaneously.
We first compute the proximity matrix $\bm{F}\in\mathbb{R}^{N}$,
\begin{equation}
    \label{F}
    \bm{F}=\tanh (\bm{t}_c\bm{W}_l\bm{t}_e^T),
\end{equation}
where $\bm{W}_l\in\mathbb{R}^{K\times K}$ is the learnable weight matrix.
The proximity matrix can be viewed as a transformation from the claim attention space to the evidence attention space.
Then we can predict the interaction attention by treating $\bm{F}$ as the feature,
\begin{equation}
    \label{Hs}
    \bm{H}^e=\tanh\big(\bm{W}_e\bm{t}_e^T+(\bm{W}_c\bm{t}_c^T)\bm{F}\big),
\end{equation}
where $\bm{W}_e$, $\bm{W}_c\in\mathbb{R}^{l\times K}$ are the learnable weight matrices.
Finally we can generate a topic similarity score between the claim and each evidence using the softmax function,
\begin{equation}
    \label{a_r}
    \bm{\alpha}^e=softmax(\bm{w}\bm{H}^e),
\end{equation}
where $\bm{w}\in\mathbb{R}^{1\times l}$ is the learnable weight,
$\bm{\alpha}^e\in\mathbb{R}^{N}$ is the attention score of each piece of evidence for the claim.
Eventually, the topic representation $\bm{A}\in \mathbb{R}^{N\times K}$ can be computed as follows,
\begin{equation}
    \label{topic content}
    \bm{A}=\bm{\alpha}^e\odot \bm{\hat{t}}_e,
\end{equation}
where $\odot$ is the dot product operation.

\paragraph{Semantic-Topic Co-attention}
\label{semantic topic information}
We weigh each piece of evidence $e_i$ to indicate the importance of the evidence and infer the claim based on
the coherence between the semantic representation and the global topic information
via the co-attention mechanism,
which is similar to the coherence-based topic attention in Section~\ref{topic information}.
More concretely,
taking $\bm{H}$ and $\bm{P}$ as input,
we compute the proximity matrix $\bm{F}\in\mathbb{R}^{K\times N}$ to transform the topic attention space to the semantic attention space by Eq.~\eqref{F}.
As a result, the attention weights $\bm{\beta}^e \in\mathbb{R}^{N}$ of evidence can be obtained by Eq. \eqref{Hs} and \eqref{a_r}.
Eventually, the semantic representation $\bm{S}\in \mathbb{R}^{N\times d}$ can be updated via $\bm{S}=\bm{\beta}^e\odot \bm{H}$.

\subsection{Aggregation Layer}
To model the implicit stances of evidence toward claim,
we incorporate the capsule network~\citep{DBLP:conf/nips/SabourFH17} into our model.
As illustrated in Figure~\ref{fig:structure},
we concatenate both the semantic representation $\bm{S}$ and
the topical representation $\bm{A}$ to form the low-level evidence capsules $\bm{u}_i=[\bm{a}_i;\bm{s}_i]\vert_{i=1}^N\in \mathbb{R}^{d_e}$.
Let $\bm{o}_j\vert_{j=1}^M \in \mathbb{R}^{d_o}$ denote the high-level class capsules,
where $M$ denotes the number of classes.
The capsule network models the relationship between the evidence capsules and the class capsules by the dynamic routing mechanism~\citep{DBLP:conf/emnlp/YangZYLZZ18},
which can be viewed as the implicit stances of each evidence toward three classes.

Formally,
let $\bm{u}_{j\vert i}$ be the predicted vector from the evidence capsule $\bm{u}_i$ to the class capsule $\bm{o}_j$,
\begin{equation}
    \label{mu}
    \bm{u}_{j\vert i}=\bm{W}_{j,i}\bm{u}_i
\end{equation}
where $\bm{W}_{j,i}\in\mathbb{R}^{d_o \times d_e}$ denotes the transformation matrix from the evidence capsule $\bm{u}_i$ to the class capsule $\bm{o}_j$.
Each class capsule aggregates all of the evidence capsules by a weighted summation over all corresponding predicted vectors:
\begin{equation}
    \begin{aligned}
        \label{vi}
        \bm{o}_j=g(\sum_{i=1}^N{\gamma_{ji}\bm{u}_{j\vert i}}), &  &
        \hat{p}_{ji}=\vert \bm{u}_i \vert,
    \end{aligned}
\end{equation}
where $g$ is a non-linear squashing function which limits the length of $\bm{o}_j$ to $[0,1]$,
$\gamma_{ji}$ is the coupling coefficient that determines the probability that the evidence capsule $\bm{u}_i$ should be coupled with the class capsule $\bm{o}_j$.
The coupling coefficient is calculated by the unsupervised and iterative dynamic routing algorithm on original logits $b_{ji}$,
which is summarized in Algorithm~\ref{algorithm:1}.
We can easily classify the claim by choosing the class capsule with the largest $\rho_j$ via the capsule loss~\citep{DBLP:conf/nips/SabourFH17}.
Moreover,
the cross entropy loss is applied on the evidence capsules to identify whether the evidence is the ground truth evidence.

\begin{algorithm}[H]\footnotesize
    \label{dynamic routing mechanism}
    \renewcommand{\algorithmicrequire}{\textbf{Procedure:}}
    \caption{Dynamic Routing Algorithm}\label{algorithm:1}
    \begin{algorithmic}[1]
        \REQUIRE{Routing($\bm{u}_{j\vert i}$, $\hat{p}_{ji}$)}
        \STATE{Initialize the logit of coupling coefficient $b_{ij}==0$; }
        \FOR{each iteration}
        \STATE{For all evidence capsule $\bm{u}_i$ and class $\bm{o}_j$: \\~~~~$\gamma_{ji}=\hat{p}_{ji}\cdot leaky\_softmax(b_{ji})$}
        \STATE{Update all the class capsules via Eq.~\eqref{vi}};
        \STATE{For all evidence capsule $\bm{u}_i$ and the class $\bm{o}_j$: \\~~~~$b_{ji}=b_{ji}+\bm{u}_{j\vert i}\cdot \bm{o}_j$}
        \ENDFOR
        \STATE{Return $\bm{o}\in\mathbb{R}^{M\times d_o}$, $\rho_j=\vert \bm{o}_j \vert_{j=1:M} $}
    \end{algorithmic}
\end{algorithm}

\section{Experimental Setting}
This section describes the datasets, evaluation metrics, baselines, and implementation details in our experiments.

\paragraph{Datasets}
We conduct experiments on two public fact checking datasets:
(1) FEVER~\citep{DBLP:conf/naacl/ThorneVCM18} is a large-scale dataset consisting of 185,455 claims along with 5,416,537 Wikipedia pages from the June 2017 Wikipedia dump.
The ground truth evidence and the label (i.e., ``SUPPORTS'', ``REFUTES'' and ``NOT ENOUGH INFO (NEI)'') are also available except in the test set.
(2) UKP Snopes~\citep{DBLP:conf/conll/HanselowskiSSLG19} is a mixed-domain dataset along with 16,508 Snopes pages.
To maintain the consistency of two datasets,
we merge the verdicts \{\emph{false}, \emph{mostly false}\}, \{\emph{true}, \emph{mostly true}\}, \{\emph{mixture}, \emph{unproven}, \emph{undetermined}\}
as \emph{``REFUTES''},\emph{``SUPPORTS''} and \emph{``NEI''}, respectively.
And we omit all other labels (i.e., \emph{legent}, \emph{outdated}, and \emph{miscaptioned}) as these instances are difficult to distinguish.
Table~\ref{statistics} presents the statistics of the two datasets.

\begin{table}[]\small
    \renewcommand\arraystretch{1.2}
    \renewcommand\tabcolsep{3.0pt}
    \centering
    \begin{tabular}{c|cccc}
        \hline
        \textbf{Datasets} & Train   & Dev    & Test   & Vocabulary size \\ \hline
        FEVER             & 145,449 & 19,998 & 19,998 & 25,753          \\
        UKP Snopes        & 4,659   & 582    & 583    & 2,258           \\ \hline
    \end{tabular}
    \caption{\label{statistics} Statistics on FEVER and UKP Snopes}
\end{table}

\paragraph{Evaluation Metrics}
The official evaluation metrics\footnote{https://github.com/sheffieldnlp/fever-scorer} for the FEVER dataset are Label Accuracy (LA) and FEVER score (F-score).
LA measures the accuracy of the predicted label $\hat{y}_i$ matching the ground truth label $y_i$ without considering the retrieved evidence.
The FEVER score labels a prediction as correct if the predicted label $\hat{y}_i$ is correct and the retrieved evidence matches at least one gold-standard evidence, 
which is a better indicator to reflect the inference capability of the model.
We use precision, recall, and macro F1 on UKP Snopes to evaluate the performance.

\paragraph{Baselines}
The following approaches are employed as the baselines,
including three top performing models on FEVER1.0 shared task (UKP Athene~\citep{DBLP:journals/corr/abs-1809-01479}, UCL MRG~\citep{yoneda2018ucl} and UNC NLP~\citep{DBLP:conf/aaai/NieCB19}),
HAN~\citep{DBLP:conf/acl/MaGJW19},
BERT-based models (SR-MRS~\citep{DBLP:conf/emnlp/NieWB19}, BERT Concat~\citep{DBLP:conf/ecir/SoleimaniMW20} and HESM~\citep{DBLP:conf/emnlp/SubramanianL20}),
and graph-based models (GEAR~\citep{DBLP:conf/acl/ZhouHYLWLS19}, Transformer-XH~\citep{DBLP:conf/iclr/ZhaoXRSBT20},
KGAT~\citep{DBLP:conf/acl/LiuXSL20} and DREAM~\citep{DBLP:conf/acl/ZhongXTXDZWY20}).

\paragraph{Implementation Details}
We describe our implementation details in this section.

\emph{Document retrieval} takes a claim along with a collection of documents as the input,
then returns $N$ most relevant documents.
For the FEVER dataset, following~\citet{DBLP:conf/coling/HanselowskiSSCC18},
we adopt the \emph{entity linking} method since the title of a Wikipedia page can be viewed as an entity and can be linked easily with the extracted entities from the claim.
For the UKP Snopes dataset, following~\citet{DBLP:conf/conll/HanselowskiSSLG19}, we adopt the \emph{tf-idf} method where the tf-idf similarity between claim and concatenation of all sentences of each Snopes page is computed,
and then the 5 highest ranked documents are taken as retrieved documents.

\emph{Evidence selection} retrieves the related sentences from retrieved documents in \emph{ranking} setting.
For the FEVER dataset, we follow the previous method from~\citet{DBLP:conf/iclr/ZhaoXRSBT20}.
Taking the concatenation of claim and each sentence as input,
the $[CLS]$ token representation is learned through BERT which is then used to learn a ranking score through a linear layer.
The hinge loss is used to optimize the BERT model.
For the UKP Snopes dataset,
we adopt the \emph{tf-idf} method from~\citet{DBLP:conf/conll/HanselowskiSSLG19},
which achieves the best precision.

\emph{Claim verification.} During the training phase,
each claim is paired with 5 pieces of evidence,
we set the batch size to 1 and the accumulate step to 8,
the layer $L$ is 3,
the head number is 5,
the $l$ is 100,
the number of class capsules $M$ is 3,
the dimension of class capsules $d_o$ is 10,
the topic number $K$ ranges from 25 to 100.
In our implementation,
the maximum length of each claim-evidence pair is 130 for both datasets.

\section{Experimental Results}
In this section, we evaluate our TARSA model in different aspects.
Firstly, we compare the overall performance between our model and the baselines.
Then we conduct an ablation study to explore the effectiveness of the topic information and the capsule network structure.
Finally, we also explore the advantages of our model in single-hop and multi-hop reasoning scenarios.

\subsection{Overall Performance}
Table~\ref{fever} and Table~\ref{snopes} report the overall performance of our model against the baselines for the FEVER dataset and the UKP Snopes dataset
\footnote{
    Note that we did not compare HESM, SR-MES and DREAM with our model on the UKP Snopes dataset for the following reasons.
    HESM requires hyperlinks to construct the evidence set, which are not available in UKP Snopes;
    SR-MRS concatenates query and context as the input to BERT, which is similar to the BERT Concat model;
    The composition of a claim in the UKP Snopes is more complicated than FEVER, which is more difficult for DERAM to construct a graph at the semantic level.}.
As shown in Table~\ref{fever},
our model significantly outperforms BERT-based models on both development and test sets.
However,
compared with the graph-based models,
TARSA outperforms previous systems, GEAR and KGAT, except DREAM for LA on the test set.
One possible reason is that DREAM constructs an evidence graph based on the semantic roles of claim and evidence,
which leverages an explicit graph-level semantic structure built from semantic roles extracted by Semantic Role Labeling~\citep{DBLP:journals/corr/abs-1904-05255} in a fine-grained setting.
Nevertheless,
TARSA shows superior performance than DREAM on the FEVER score,
which is a more desirable indicator to demonstrate the reasoning capability of the model.
As shown in Table~\ref{snopes},
TARSA performs the best compared with all previous approaches on the UKP Snopes dataset.

\begin{table}[]\footnotesize
    \renewcommand\arraystretch{1.2}
    \renewcommand\tabcolsep{3.5pt}
    \centering
    \begin{tabular}{lcccc}
        \hline
        \multicolumn{1}{c}{\multirow{3}{*}{\textbf{Models}}} & \multicolumn{4}{c}{\textbf{FEVER}}                                                                    \\ \cline{2-5}
        \multicolumn{1}{c}{}                                 & \multicolumn{2}{c}{Dev}            & \multicolumn{2}{c}{Test}                                         \\ \cline{2-5}
        \multicolumn{1}{c}{}                                 & LA                                 & F-score                  & LA                & F-score           \\ \hline
        UKP Athene                                           & 68.49                              & 64.74                    & 65.46             & 61.58             \\
        UCL MRG                                              & 69.66                              & 65.41                    & 67.62             & 62.52             \\
        UNC NLP                                              & 69.72                              & 66.49                    & 68.21             & 64.21             \\ \hline
        HAN                                                  & 72.00                              & 57.10                    & -                 & -                 \\
        BERT(base)                                           & 73.51                              & 71.38                    & 70.67             & 68.50             \\
        BERT(large)                                          & 74.59                              & 72.42                    & 71.86             & 69.66             \\
        BERT Pair                                            & 73.30                              & 68.90                    & 69.75             & 65.18             \\
        BERT Concat                                          & 73.67                              & 68.89                    & 71.01             & 65.64             \\
        SR-MRS                                               & 75.12                              & 70.18                    & 72.56             & 67.26             \\
        HESM(ALBERT Base)                                    & 75.77                              & 73.44                    & 73.25             & 70.06             \\ \hline
        GEAR                                                 & 74.84                              & 70.69                    & 71.60             & 67.10             \\
        KGAT(BERT base)                                      & 78.02                              & 75.88                    & 72.81             & 69.40             \\
        KGAT(BERT large)                                     & 77.91                              & 75.86                    & 73.61             & 70.24             \\
        DREAM                                                & \underline{79.16}                  & -                        & \textbf{76.85}    & \underline{70.60} \\
        Transformer-XH                                       & 78.05                              & 74.98                    & 72.39             & 69.07             \\
        \textit{our} TARSA                                   & \textbf{81.24}                     & \textbf{77.96}           & \underline{73.97} & \textbf{70.70}    \\ \hline
    \end{tabular}
    \caption{\label{fever} Overall performance on the FEVER dataset (\%).}
\end{table}

\begin{table}[]\small
    \renewcommand\arraystretch{1.2}
    \centering
    \begin{tabular}{lccc}
        \hline
        \multicolumn{1}{c}{\multirow{2}{*}{\textbf{Models}}} & \multicolumn{3}{c}{\textbf{UKP Snopes}}                                         \\ \cline{2-4}
        \multicolumn{1}{c}{}                                 & Precision                               & Recall            & macro F1          \\ \hline
        Random baseline                                      & 0.333                                   & 0.333             & 0.333             \\
        Majority vote                                        & 0.170                                   & 0.198             & 0.249             \\ \hline
        BERTEmb                                              & 0.493                                   & 0.477             & 0.485             \\
        BERT Concat                                          & 0.485                                   & 0.474             & 0.478             \\ \hline
        GEAR                                                 & 0.368                                   & 0.337             & 0.352             \\
        KGAT                                                 & 0.493                                   & 0.440             & 0.465             \\
        Transformer-XH                                       & \underline{0.532}                       & \underline{0.529} & \underline{0.531} \\
        \emph{ours} TARSA                                    & \textbf{0.611}                          & \textbf{0.540}    & \textbf{0.573}    \\ \hline
    \end{tabular}
    \caption{\label{snopes} Overall performance on the UKP Snopes dataset.}
\end{table}

\subsection{Effect of Topic Number}
Table~\ref{topic number} shows the results of our TARSA model with different number of topics on the development set of FEVER and UKP Snopes.
It can be observed that the optimal topic number is 25 for FEVER and 50 for UKP Snopes.
One possible reason is that UKP Snopes is retrieved from multiple domains which includes more diverse categories than those of FEVER.
\begin{table}[]\footnotesize
    \renewcommand\arraystretch{1.2}
    \centering
    \begin{tabular}{ccc|ccc}
        \hline
        \multirow{2}{*}{\textbf{\#Topic}} & \multicolumn{2}{c|}{\textbf{FEVER (\%)}} & \multicolumn{3}{c}{\textbf{UKP Snopes}}                                                    \\ \cline{2-6}
                                          & LA                                       & F-score                                 & P.             & R.             & macro F1       \\ \hline
        25                                & \textbf{81.24}                           & \textbf{77.96}                          & 0.560          & 0.539          & 0.549          \\
        50                                & 80.30                                    & 77.13                                   & \textbf{0.611} & 0.540          & \textbf{0.573} \\
        75                                & 80.62                                    & 77.38                                   & 0.563          & \textbf{0.564} & 0.564          \\
        100                               & 80.30                                    & 77.13                                   & 0.592          & 0.533          & 0.561          \\ \hline
    \end{tabular}
    \caption{\label{topic number} Evaluation of TARSA with different number of topics on FEVER and UKP Snopes.}
\end{table}

\subsection{Ablation Study}
To further illustrate the effectiveness of the topic information and the capsule-level aggregation modeling,
we perform an ablation study on the development set of FEVER.

\paragraph{Effect of Topic Information:}
We first explore how the model performance is impacted by the removal of various topic components.
The first six rows in Table~\ref{ablation} present the label accuracy (LA) and the FEVER score on the development set of FEVER after removing various components,
where \emph{STI} denotes the semantic-topic information in Section~\ref{semantic topic information},
${TC}_{ee}$ denotes the topical coherence among multiple pieces of evidence,
$TC_{ce}$ denotes the topical consistency between the claim and each piece of evidence.
As expected,
LA and the FEVER score decrease consistently with a gradual removal of various components,
which demonstrates the effectiveness of incorporating topic information in three aspects.
We find that after all modules are removed,
the performance of TARSA is still nearly 2\% higher than our base model, Transformer-XH, due to the use of the capsule network in TARSA.

\paragraph{Effect of Capsule-level Aggregation:}
We explore the effectiveness of the capsule-level aggregation by comparing it with four different aggregation methods.
The last four rows in Table~\ref{ablation} show the results of aggregation analysis in the development set on FEVER.
The max pooling, sum, and mean aggregation consider the learned representations of evidence as a single matrix,
then apply a linear layer to classify the input claim as SUPPORTS, REFUTES, or NEI.
The attention-based aggregation method is used in~\citet{DBLP:conf/acl/ZhouHYLWLS19},
where the dot-product attention is computed between the claim and each evidence to weigh them differently.
Finally, our TARSA model aggregates the information of all pieces of evidence using the capsule network,
which connects the evidence capsules to the class capsules in a clustered way.
From the results,
our model outperforms all other aggregation methods.

\begin{table}[]\footnotesize
    \renewcommand\arraystretch{1.2}
    \centering
    \begin{tabular}{llcc}
        \hline
        \multicolumn{2}{c}{\textbf{Models}}                        & LA              & F-score         \\ \hline
        \multicolumn{2}{l}{\emph{our} TARSA}                       & 81.24           & 77.96           \\
        \multicolumn{2}{l}{-\emph{STI}}                            & 80.62           & 77.38           \\
        \multicolumn{2}{l}{-${TC}_{ee}$}                           & 80.51           & 77.31           \\
        \multicolumn{2}{l}{-$TC_{ce}$}                             & 80.35           & 77.16           \\
        \multicolumn{2}{l}{-$TC_{ee}$ - $TC_{ce}$}                 & 80.06           & 76.88           \\
        \multicolumn{2}{l}{-$TC_{ee}$ - $TC_{ce}$ - \emph{STI}}    & 79.93           & 76.80           \\ \hline \hline
        \multicolumn{1}{l|}{\multirow{4}{*}{\textbf{Aggregation}}} & max pooling     & 79.36   & 76.33 \\
        \multicolumn{1}{l|}{}                                      & sum             & 79.60   & 76.57 \\
        \multicolumn{1}{l|}{}                                      & mean            & 79.28   & 76.19 \\
        \multicolumn{1}{l|}{}                                      & attention-based & 79.52   & 76.45 \\  \hline
    \end{tabular}
    \caption{\label{ablation} Ablation analysis in the development set of FEVER.}
\end{table}

\begin{table}[]\footnotesize
    \renewcommand\arraystretch{1.1}
    \centering
    \begin{tabular}{lcccc}
        \hline
        \multicolumn{1}{c}{\multirow{2}{*}{\textbf{Models}}} & \multicolumn{2}{c}{\textbf{Single-hop}} & \multicolumn{2}{c}{\textbf{Multi-hop}}                                   \\ \cline{2-5}
        \multicolumn{1}{c}{}                                 & LA                                      & F-score                                & LA             & F-score        \\ \hline
        BERT Concat                                          & 89.93                                   & 84.23                                  & 92.74          & 89.92          \\
        GEAR                                                 & 81.56                                   & 76.62                                  & 89.21          & 86.66          \\
        KGAT                                                 & 90.99                                   & 85.22                                  & 93.73          & 90.93          \\
        Transformer-XH                                       & 89.23                                   & 83.50                                  & 93.39          & 90.71          \\
        \emph{our} TARSA                                     & \textbf{91.30}                          & \textbf{85.48}                         & \textbf{94.82} & \textbf{92.03} \\ \hline
    \end{tabular}
    \caption{\label{single and multi} Fact verification accuracy on claims that require Single and Multiple pieces of evidence.}
\end{table}

\subsection{Performance on Different Scenarios}
Table~\ref{single and multi} presents the performance of our model on single-hop and multi-hop reasoning scenarios on the FEVER dataset compared with several baselines.
The single-hop mainly focuses on the denoising ability of the model with the retrieved evidence,
which selects the salient evidence for inference.
The multi-hop mainly emphasizes the relatedness of different pieces of evidence for the joint reasoning,
which is a more complex task.

We build the training and testing sets for both single-hop and multi-hop scenarios based on the number of gold-standard evidence of a claim.
If more than one gold-standard evidence is required,
then the claim would require multi-hop reasoning.
The instances with the NEI label are removed because there is no gold-standard evidence matching this label.
The single-hop reasoning set contains 78,838 and 9,682 instances for training and testing, respectively,
while the multi-hop reasoning set contains 30,972 and 3,650 instances for training and testing, respectively.
As Table~\ref{single and multi} shows,
TARSA outperforms all other baselines on LA by at least 0.31\% in the single-hop scenario and 1.09\% in the multi-hop scenario, respectively,
which shows a consistent improvement in both scenarios.
In addition,
TARSA is more effective on the multi-hop scenario as the capsule-level aggregation helps better aggregate the information of all pieces of evidence.

\begin{table}[]\small
    \renewcommand\arraystretch{1.3}
    \renewcommand\tabcolsep{5.0pt}
    \centering
    \begin{tabular}[c]{c|p{5cm}}
        \hline
        \multicolumn{2}{c}{\textbf{Example}: REFUTES }                                                                                                                                                                                                                                                                                                                                                               \\ \hline
        \multicolumn{1}{c|}{\textbf{Claim}}                           &
        During an \textcolor{blue}{interview} with the \textcolor{blue}{Washington} Post, \textcolor{blue}{President} Obama stated that \textcolor{blue}{Americans} would be better off under \textcolor{blue}{martial} \textcolor{blue}{law}.                                                                                                                                                                       \\ \hline
        \multicolumn{1}{c|}{\multirow{5}{*}[-1ex]{\textbf{Evidence}}} &
        \textbf{e1}: In a \textcolor{blue}{statement} appearing in the \textcolor{blue}{Washington} Post, \textcolor{blue}{United States President} Barrack Hussein Obama said \textcolor{blue}{Americans} would be better living under \textcolor{blue}{martial law}.                                                                                                                                               \\ \cline{2-2}
        \multicolumn{1}{c|}{}                                         &
        \textbf{e2}: The \textcolor{blue}{Washington} Post, a long time \textcolor{blue}{democratic} \textcolor{blue}{mouth} piece and Obama supporter, downplayed the \textcolor{blue}{statement} by suggesting it was made in jest and that \textcolor{blue}{President} Obama had been joking around'' with the \textcolor{blue}{reporter} at the \textcolor{blue}{time} the \textcolor{blue}{statement} was made. \\ \cline{2-2}
        \multicolumn{1}{c|}{}                                         &
        \textbf{e3}: A \textcolor{blue}{Washington} insider, speaking under conditions of anonymity, reveals that Obama made additional inflammatory \textcolor{blue}{comments} not \textcolor{blue}{reported} by the \textcolor{blue}{Washington} Post.                                                                                                                                                             \\ \cline{2-2}
        \multicolumn{1}{c|}{}                                         &
        \textbf{e4}: \textcolor{blue}{Americans} have had their \textcolor{blue}{chance} to aspire to be better, to rise to the \textcolor{blue}{occasion}, but \textcolor{blue}{time} and again they \textcolor{blue}{fail}.                                                                                                                                                                                        \\ \cline{2-2}
        \multicolumn{1}{c|}{}                                         &
        \textbf{e5}: Would tighter \textcolor{blue}{restrictions} really be such an imposition?                                                                                                                                                                                                                                                                                                                      \\ \hline
    \end{tabular}
    \caption{\label{case study} Example of retrieved evidence ranked by the topical consistency between the claim and each piece of evidence. The topic words are marked in blue.}
\end{table}

\subsection{Case Study}
Table~\ref{case study} illustrates an example from the UKP Snopes dataset which is correctly detected as \emph{REFUTES},
where the topic words extracted by LDA are marked in blue.
From the table we can observe:
1) the top two pieces of evidence (i.e., $e1$ and $e2$) have higher topical overlap with the claim and also with each other;
2) the lower two pieces of evidence (i.e., $e4$ and $e5$) seem less important because they are less topically relevant to the claim;
3) for $e3$, it is difficult to judge its relevance from either the topical or the semantic perspective,
which is ambiguous for the identification of the truthfulness of the claim.

\subsection{Error Analysis}
We randomly select 100 incorrectly predicted instances from FEVER and UKP Snopes datasets and categorize the main errors.
The first type of errors is caused by the quality of topics extracted by LDA.
This is because the average length of sentences in both datasets is much shorter after removing the low- and high-frequency tokens,
which poses a challenge for LDA to extract high quality topics to match the topical consistency between a claim and each evidence.
The second type of errors is due to the failure of detecting multiple entity mentions referring to the same entity.
For example, the claim describes ``\emph{Go Ask Alice was the real life diary of a teenager girl}'', where evidence describes that ``\emph{This book is a work of fiction}''.
The model fail to understand the relationship between \emph{diary} and \emph{fiction}.

\section{Conclusion}
We have presented a novel topic-aware evidence reasoning and stance-aware aggregation model for fact verification.
Our model jointly exploits the topical consistency and the semantic interaction to learn evidence representations at the sentence level.
Moreover,
we have proposed the use of the capsule network to model the implicit stances of evidence toward a claim for a better aggregation of information encoded in evidence.
The results on two public datasets demonstrate the effectiveness of our model.
In the future,
we plan to explore an iterative reasoning mechanism for more efficient evidence aggregation for fact checking.


\section*{Acknowledgements}
We would like to thank anonymous reviewers for their valuable comments and helpful suggestions.
This work was funded by the National Key Research and Development Program of China (2016YFC1306704),
the National Natural Science Foundation of China (61772132),
and the EPSRC (grant no. EP/T017112/1, EP/V048597/1).
YH is supported by a Turing AI Fellowship funded by the UK Research and Innovation (UKRI) (grant no. EP/V020579/1).

\bibliographystyle{acl_natbib}
\bibliography{anthology,acl2021}


\end{document}